\documentclass[10pt,twocolumn,letterpaper]{article}

\usepackage[pagenumbers]{cvpr} 

\usepackage{lipsum}  
\usepackage{graphicx}
\usepackage{amsmath}
\usepackage{amssymb}
\usepackage{multirow}
\usepackage{enumitem}
\usepackage{balance}
\usepackage{adjustbox}
\usepackage{array,ragged2e}
\usepackage[table]{xcolor}

\usepackage{algorithmic}
\usepackage[linesnumbered,ruled]{algorithm2e}
\SetAlgorithmName{Algorithm}{Algorithm}{Algorithm}
\IncMargin{0.5em}
\SetCommentSty{textnormal}
\SetNlSty{}{}{:}
\SetAlgoNlRelativeSize{0}
\SetKwInput{KwGlobal}{Global}
\SetKwInput{KwPrecondition}{Precondition}
\SetKwProg{Proc}{Procedure}{:}{}
\SetKwProg{Func}{Function}{:}{}
\SetKw{And}{and}
\SetKw{Or}{or}
\SetKw{To}{to}
\SetKw{DownTo}{downto}
\SetKw{Break}{break}
\SetKw{Continue}{continue}
\SetKw{SuchThat}{\textit{s.t.}}
\SetKw{WithRespectTo}{\textit{wrt}}
\SetKw{Iff}{\textit{iff.}}
\SetKw{MaxOf}{\textit{max of}}
\SetKw{MinOf}{\textit{min of}}
\SetKwBlock{Match}{match}{}{}

\usepackage{floatrow}
\newfloatcommand{capbtabbox}{table}[][\FBwidth]

\usepackage{wrapfig}

\usepackage{silence}

\makeatletter
\DeclareRobustCommand{\iscircle}{\mathord{\mathpalette\is@circle\relax}}
\newcommand\is@circle[2]{%
  \begingroup
  \sbox\z@{\raisebox{\depth}{$\m@th#1\bigcirc$}}%
  \sbox\tw@{$#1\square$}%
  \resizebox{!}{\ht\tw@}{\usebox{\z@}}%
  \endgroup
}
\makeatother

\makeatletter
\newcommand{\xRightarrow}[2][]{\ext@arrow 0359\Rightarrowfill@{#1}{#2}}
\makeatother

\pdfoutput=1

\definecolor{cvprblue}{rgb}{0.21,0.49,0.74}
\usepackage[pagebackref,breaklinks,colorlinks,allcolors=cvprblue]{hyperref}

\def\paperID{12250} 
\def\confName{CVPR}
\def\confYear{2025}

\title{VORD: Visual Ordinal Calibration for Mitigating Object Hallucinations \\ in Large Vision-Language Models}

\author{Dexter Neo, Tsuhan Chen\\
School of Computing\\
National University of Singapore\\
{\tt\small e0534450@u.nus.edu,tsuhan@nus.edu.sg}
}

\begin{document}
\maketitle

\begin{abstract}
Large Vision-Language Models (LVLMs) have made remarkable developments along with the recent surge of large language models. Despite their advancements, LVLMs have a tendency to generate plausible yet inaccurate or inconsistent information based on the provided source content. This phenomenon, also known as ``hallucinations" can have serious downstream implications during the deployment of LVLMs. To address this, we present VORD a simple and effective method that alleviates hallucinations by calibrating token predictions based on ordinal relationships between modified image pairs. VORD is presented in two forms: 1.) a minimalist training-free variant which eliminates implausible tokens from modified image pairs, and 2.) a trainable objective function that penalizes unlikely tokens. Our experiments demonstrate that VORD delivers better calibration and effectively mitigates object hallucinations on a wide-range of LVLM benchmarks. Our code is available at: \textit{\href{https://github.com/dexterdley/VORD}{https://github.com/dexterdley/VORD} }.
\end{abstract}
\section{Introduction}
\label{sec:intro}
Vital requirements for the large-scale adoption of Large Vision-Language Models (LVLMs) include their correctness and faithfulness of generated content. Although LVLMs have achieved significant success in performing complex tasks such as image captioning \cite{Hu_2022_CVPR} and visual question answering \cite{Zhang_2024_CVPR}, they are prone to ``hallucinations". Specifically, LVLMs tend to have \textbf{object hallucinations} (OH) - instances where the model produces plausible yet incorrect descriptions that are inconsistent to the given visual context cues.

Hallucinations remain a major obstacle for the deployment of LVLMs in high-stakes, risk-sensitive applications, such as healthcare \citep{wang2023chatcad, yildirim2024multimodal}, autonomous agents \cite{langsuite2023, chen2024drivingwithllms, Wei_2024_CVPR, ma2023dolphins} and legal AI \citep{wu2023precedentenhanced, NEURIPS2023_91f18a12}, where they can result in unsafe or undesirable outcomes. These downstream tasks require LVLMs to not only be accurate but also confident about their predictions. Making it essential for LVLMs to be able to communicate their uncertainty in the event of OH. 

\begin{figure}[!tb]
\centering
\includegraphics[width=\columnwidth]{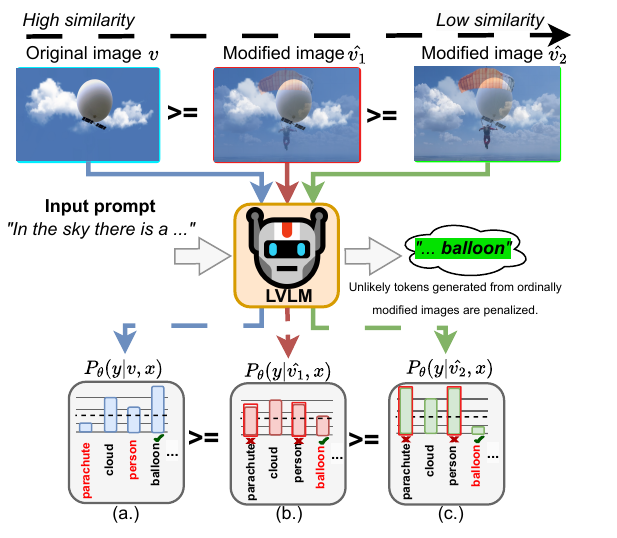}
\caption{VORD suppresses hallucinated objects such as \textcolor{red}{\texttt{<person>}} by enforcing ordinal ranking of confidences and penalizing unlikely tokens during the generation process.}
\label{fig:fig1}
\end{figure}

The challenge behind mitigating OH comes from the underlying fundamental issues with visual-question answering benchmarks, such as the over-dependence of the natural language priors embedded in the backbone large language model and the prevalent statistical imbalances in the training corpus \cite{VCD_2024_CVPR, M3ID_2024_CVPR}. \cref{fig:fig1} illustrates this phenomenon, where incorrect tokens such as $\texttt{<person>}$ and $\texttt{<parachute>}$ can have non-zero probabilities, leaving them susceptible to being mistakenly sampled during generation, resulting in object hallucinations.

While numerous efforts have been placed on how to reduce OH in LVLMs \cite{Yin2023WoodpeckerHC, LURE24, VCD_2024_CVPR, M3ID_2024_CVPR, OPERA_2024_CVPR}. Current studies tend to focus on improving the overall accuracy of LVLMs, leaving the critical aspect of model confidence calibration in OH largely unaddressed. Since model calibration seeks to align a model's confidence with its correctness, well-calibrated token generation could also directly lead to better text generations in LVLMs.

Recent works \cite{pmlr-v235-tu24a} have shown that visual language models such as CLIP \cite{clip-radford21a} are innately better calibrated than other models trained on ImageNet \cite{NEURIPS2021_8420d359, galil2023what}. While other studies \cite{NEURIPS2023_2c6be9f0, ChatGPT4_report} have suggested that calibration can be significantly reduced after additional fine-tuning, due to the quality and distribution of the training corpus. Following this line of research, our goal is to reduce object hallucinations, while improving LVLM calibration.

In this paper, we first show that when left unchecked, LVLMs tend to have tokenwise probabilities that do not conform to ranked ordinal relationships when faced with \textit{visually corrupted} inputs (see \cref{sec:motivation}). Building upon this observation, we propose \textbf{V}isual \textbf{Ord}inal (VORD) calibration for mitigating object hallucinations in LVLMs. VORD can be implemented as a lightweight, minimalist \textit{training-free} decoding strategy or used as a fully optimizable \textit{objective function} if training resources are available. We demonstrate that VORD effectively mitigates object hallucinations in LVLMs, along with better calibrated predictions and shorter generated text sequences on a wide plethora of object hallucination benchmarks. Our main contributions are summarized as follows:
\begin{itemize}[noitemsep,topsep=0pt]
\item \textbf{Visually Ordinal Tokens} We demonstrate that when faced with corrupted visual cues, LVLMs do not exhibit well-ordered token probability distributions.
\item \textbf{VORD Decoding:} We present VORD Decoding, a lightweight, minimalist and training-free method for contrastive decoding in LVLMs.
\item \textbf{VORD Loss:} We also present VORD Loss, an ordinal ranking loss function for improving performance and mitigating hallucinations in LVLMs.
\item \textbf{Adaptive Visual Similarity Margin:} Additionally, we enforce the transitive properties of VORD with a novel dynamic visual-similarity penalty margin
\end{itemize}

\section{Related Work}
\paragraph{Visual Language Models}
Given the recent influx of large language models (LLMs), LVLMs have demonstrated incredible capabilities across a multitude of applications. Initial works on VLMs include BERT-based text decoders \cite{devlin-etal-2019-bert, liu2019roberta} that integrate textual and visual inputs \cite{Sun_2019_ICCV, li-etal-2020-bert-vision, BLIP_icml22, wang2022git}. Presently, a large focus of LVLM research pivots on Visual Instruction Fine-tuning \cite{LLaVA_neurips23, LLaVA_2024_CVPR, VILA_2024_CVPR} which extends instruction tuning from NLP research to allow LVLMs to learn on text-image pairs with instructions or descriptions of the image. This process allows backbone LLMs such as LLaMA \citep{LLAMA_2023} to better ``see" visual tokens and understand the context, delivering better performance on a wide range of tasks such as image captioning and visual question answering.

\paragraph{Hallucinations in LVLMs}
Despite their incredible advancements, LVLMs are susceptible to generating plausible-sounding yet ungrounded falsehoods \cite{M3ID_2024_CVPR}. This phenomena also widely referred to as ``hallucinations" \cite{zhou-etal-2021-detecting, xu-etal-2023-understanding, truthfulqa-2022, Zhang2023SirensSI, shi-etal-2024-replug} have mystified researchers in both the NLP and CV communities \cite{Kalai2023CalibratedLM}. Various approaches have been proposed on how to better reduce OH in LVLMs, all closely related to our work. These include self-correction and self-questioning techniques \cite{Yin2023WoodpeckerHC, LURE24, sun2024sq} that ground the LVLM to object/statistical evidence. Other works include training-free decoding approaches that aim to alleviate OH by either reducing/penalizing reliance on textual priors \citep{VCD_2024_CVPR, M3ID_2024_CVPR, OPERA_2024_CVPR, HALC_icml24}. Parallel to this line of research, our work seeks to better analyze hallucinations from a confidence calibration standpoint.

\paragraph{Deep Neural Network Calibration} 
Calibration aligns a model's correctness with its confidence estimates. Relevant examples of different calibration methods include; (a) Confidence penalizers that restrict model confidences \citep{Mller2019WhenDL, moon2020crl, Hebbalaguppe_2022_CVPR, liu2022mbls, Cheng_2022_CVPR, liu2023cals, Liu_2023_CVPR}. (b) Regularizers that interpolate the model's loss optimization space \citep{Gal2015DropoutAA, zhang2018mixup, sapkota_neurips2023, Noh_2023_ICCV}. For this work, VORD draws inspiration from the confidence penalty-regularizer techniques in (a) and (b) to deliver well-calibrated ordinal LVLM predictions.

\section{Background and Motivation}
\subsection{Preliminaries}
We begin by considering a decoding problem for a LVLM parameterized by $\theta$. Given an input textual prompt $x$ and input visual context $v$, the LVLM is simply a mapping $P_{\boldsymbol{\theta}}: X, V \rightarrow Y$ between the input vocabulary \& visual space $(X, V)$ to the predicted output token space $Y$. \cref{fig:vord_overview} illustrates a typical LVLM architecture, which mainly consists of three major components; a \textit{vision encoder}, such as a vision transformer (ViT) denoted by $f_\theta$, a \textit{LLM} and a \textit{projector} that connects the visual and text tokens from the two models. The goal of the LVLM is to autoregressively generate an appropriate response $y$ sampled from the probability distribution conditioned on the textual query and visual context. Specifically:
\begin{equation}
\begin{split}
&y_t \sim P_\theta(y_t| v, x, y_{<t}) \propto \exp h_\theta(y_t| v, x, y_{<t})
\end{split}
\end{equation}
where $y_t$ is the generated token at time step $t$, $y_{<t}$ represents the sequence of sampled tokens preceding $y_t$ and $h_\theta$ are the logits obtained from the penultimate layer. The conditional probabilities of the model are obtained after the $\mathrm{softmax}$ function, which are then used to train the LVLM by maximizing the likelihoods of a valid sequence conditioned on the visual context and text prompt: $\max_\theta \Pi_{t=1}^T P_\theta(y_t|v, x, y_{<t})$. We consider a model perfectly calibrated if $\mathbb{P}(\hat{y}=y| \hat{P}=P ) = P \quad \forall \in P[0-1]$, whereby the predicted token confidence accurately reflects the probability of that token being correct. To measure calibration error, the commonly used metric is the expected calibration error (ECE) \citep{10.5555/2888116.2888120}:
\begin{equation}
    \text{ECE} = \sum^B_{b=1} \frac{n_b}{N} | \textit{acc}(b) - \textit{conf}(b)|.
\end{equation}
which divides predictions into $B$ bins of $n_b$ samples, measuring the weighted absolute difference between the average correctness $\textit{acc}(b)$ and confidences $\textit{conf}(b)$ of each bin. 

A key challenge in calibrating LVLMs stems from the scarcity of target sequences. In ideal scenarios, we would have multiple examples of correct sequences for each input context and prompt, allowing the model to calibrate the assigned confidences of each token \cite{zhao2023calibrating}. Unfortunately, most visual question answering training datasets only have a single target sequence per input, making it difficult to obtain well-calibrated predictions. Recent studies \cite{liu-liu-2021-simcls, liu-etal-2022-brio, ravaut-etal-2022-summareranker, zhao2023calibrating} have proposed two-stage methods that involves the generation and re-ranking of token candidates for LLMs. Our work closely aligns with this line of research, but from a vision-based perspective for LVLMs.

\begin{figure}[!tb]
    \centering
    \small
    \includegraphics[width=\columnwidth]{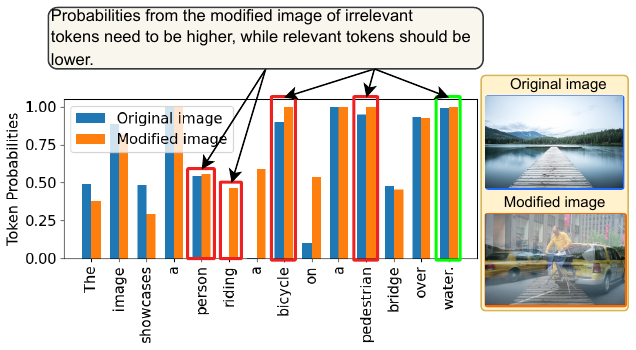}
    \caption{Comparisons between tokens probabilities obtained from the original and modified image. We observe a non-ordinal relations between both sets of probability distributions. The prompt used is: ``Describe this image in detail." }
    \label{fig:vord_tokens}
\end{figure}

\subsection{Why do LVLMs Hallucinate?}
Object hallucinations in LVLMs can largely be attributed to various reasons, such as: (1) Statistical imbalances and object correlations of the training corpus \cite{agrawal-etal-2016-analyzing, Goyal_2017_CVPR, Agarwal_2020_CVPR, Biten_2022_WACV, LURE24} which can cause the model to generate references to objects with higher occurrences in the training data. (2) Intrinsic biases and over-reliance on textual priors ingrained within the large language model \cite{pope-2023-evaluating, Ye2023mPLUGOwlME, VCD_2024_CVPR, M3ID_2024_CVPR}, resulting in the model prioritizing textual consistency as opposed to factual consistency. (3) Model's inability to accurately differentiate context from facts during the generation process \cite{HALC_icml24}, which can lead to the inclusion of irrelevant or erroneous details. Recent studies indicate that hallucinations are not random, but follow certain patterns and tendencies such as the propensity to generate common objects or those closely related to the generated text \cite{LURE24, OPERA_2024_CVPR}. 

\begin{figure}[!tb]
    \centering
    \includegraphics[width=\columnwidth]{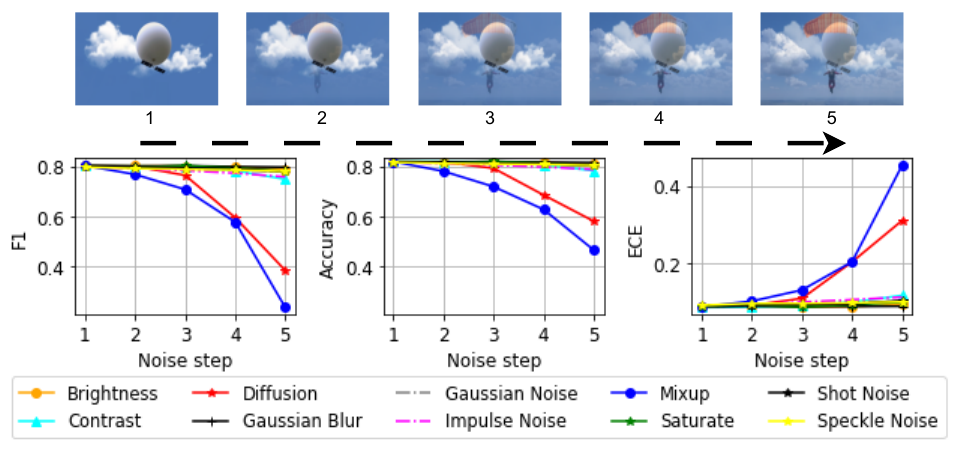}
    \vspace{-2.0em}
    \caption{Visual corruptions, such as random noise and image mixing can introduce uncertainty into LVLMs. Our findings indicate that Mixup can be a particularly effective technique for inducing uncertainty, leading to more significant errors than diffusion noise.}
    \label{fig:vord_mixup}
\end{figure}

\subsection{Inducing Visual Uncertainties in LVLMs}
The visual quality of images fed into LVLMs directly impacts their generative abilities and faithfulness. If these input images are corrupted, it can lead to errors in the model's output. Recent studies have shown that by inducing visual uncertainties \eg Diffusion noise \cite{VCD_2024_CVPR, AGLA_2024} into the input images, LVLMs are more likely to generate incorrect/hallucinatory descriptions. 

These visual uncertainties can cause the model to develop a propensity to generate common objects that frequently appear in the training data \cite{pope-2023-evaluating, LURE24}, neglecting important visual context cues. In other words, LVLMs may prioritize generating familiar or expected textual content over accurately representing the visual context, especially when confronted with visual uncertainty. \cref{fig:vord_mixup} shows the sensitivity of LVLMs to different visual corruption types such as Contrast and Brightness \cite{hendrycks2019robustness}, with Diffusion noise and Mixup causing significant performance drops. While a detailed analysis of different corruption types is beyond the scope of this work, we include additional ablation studies on different corruptions in \cref{appendix:additional_ablation_studies}.

\subsection{Do LVLMs Hallucinate Ordinally?}
\label{sec:motivation} 
Although recent studies have explored the effects of visual uncertainties on LVLMs and how contrasting generated tokens can help alleviate hallucinations. An interesting question remains: Do LVLMs hallucinate ordinally under visual uncertainty? In other words, do the conditional probabilities of generated tokens consistently follow a ranked order based on modifications to the given input?

To investigate this, we prompt a LVLM on different levels of modifications applied to the original image. These modified images are presented to the LVLM to generate new and irrelevant outputs. An intriguing observation illustrated in \cref{fig:vord_tokens}, was that some tokens generated from the modified image $\hat{v}$ do not always obey the expected transitive property, suggesting that LVLMs do not consistently hallucinate in a ranked ordinal pattern based on visual uncertainties. Specifically, tokens of interest from the clean image should ideally have higher probabilities $P_\theta(y_t|v,x,y_{<t})$ than those from the modified image $P_\theta(y_t|\hat{v}, x, y_{<t})$\footnote{~For convenience in notation, we abbreviate $P_\theta(y_t|v,x,y_{<t})$ as $P_\theta(y_t|v,x)$ in the rest of this paper.}. For the simple case with Mixup \cite{zhang2018mixup}, we expect the following relationship to hold:
\begin{equation}
    1.0 \ge \lambda_i \ge \lambda_j \Leftrightarrow P_\theta(y_t|v,x) \ge P_\theta(y_t|\hat{v}, x)
    \label{eq:transitive}
\end{equation}
Where $\lambda \in [0, 1] \sim \text{Beta}(\alpha, \alpha)$ is a hyperparameter randomly drawn from a Beta distribution and the modified sample obtained as: $\hat{v} = \lambda  v_i + (1 - \lambda) v_j$. Our findings demonstrate that under the influence of Mixup, while undesirable tokens (in red) such as $\texttt{<person>}$ and $\texttt{<bicycle>}$ have higher probabilities than the original, making them easy to identify as hallucinations. On the other hand, desired tokens (in green) such as $\texttt{<water>}$ should have much lower probabilities after modifications and did not behave as expected. To address this, we introduce VORD in the upcoming sections and also look at other variants of corruptions that can be used to induce visual uncertainties.
\begin{figure*}[!tb]
\centering
\includegraphics[width=0.9\textwidth]{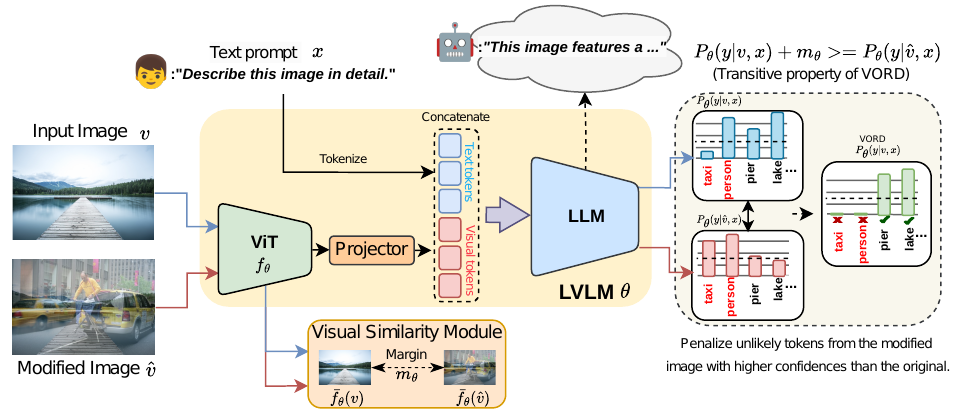}
\caption{VORD penalizes tokens with higher conditional probabilities from $\hat{v}$ than those from the original in $v$. The enforcement of the transitive property in VORD helps improve model performance, by filtering unlikely tokens during generation and training.} 
\label{fig:vord_overview}
\end{figure*}

\section{VORD: Visual Ordinal Calibration}
Our findings in \cref{sec:motivation} demonstrate that under the influence of image corruptions, LVLMs do not follow a consistent relationship of token confidences. Leveraging off of this finding, we propose Visual Ordinal Calibration for calibrating LVLM text generations based on ordinal visual cues. VORD is designed to handle the negative influences of statistical biases and textual priors that lead to object hallucinations, while calibrating tokenwise confidences in an ordinal fashion. This section introduces VORD in two forms: 1.) a lightweight and cost-effective decoding variant and 2.) a learnable objective function for finetuning LVLMs. Lastly, we discuss our adaptive penalty margin used in both forms of VORD, based on the visual similarities between the original and modified image pairs.

\subsection{Design 1: VORD Decoding}
\subsubsection{Visually Ordinal Tokens}
The key principles of the transitive property in VORD can be utilized into a training-free decoding method. Given an original input image $v$, modified image $\hat{v}$ and text prompt $x$, the model generates two sets of conditional probabilities based off $P_\theta(y_t|v, x, y_{<t})$ and $P_\theta(y_t|\hat{v}, x, y_{<t})$. We only consider tokens that have a higher probability than their counterparts generated from the modified image $\hat{v}$. VORD decoding is formalized as follows:
\begin{equation}
    P^\text{VORD}_\theta(y|v,\hat{v},x) =
    \begin{cases}
    P_\theta(y_t|v, x), & \text{accept, if } \delta_\theta \\
    0, & \text{otherwise reject}
    \end{cases}
    \label{eq:vord_decoding}
\end{equation}
where $\delta_\theta = P_\theta(y_t|v, x) + m_\theta  \ge P_\theta(y_t|\hat{v}, x)$ is the ordinal mask that accepts/rejects tokens based on the transitive property of VORD, with the margin $m_\theta \in [0 - 1]$ controlling the stringency of acceptance. This condition ensures that the model prioritizes tokens that are consistent to the original image, mitigating the introduction of irrelevant or hallucinated elements. In scenarios where the modified image is visually equivalent to the original, the ordinal condition in \cref{eq:vord_decoding} is not activated and VORD reduces to regular decoding. Subsequently, additional sampling strategies such as beam search \cite{beam-search-2017}, top-k sampling \cite{top-k-2018-fan-etal} and nucleus sampling \cite{nucleus_sampling_Holtzman2020} can also be performed together with VORD.

\begin{algorithm}[!tb]
\SetAlgoNoLine
\SetAlgoNoEnd
\DontPrintSemicolon
\KwData{Given training corpus $D_\text{train} = (x_i, y_i)^N_{i=1}$ }
\SetKwFunction{fOptimize}{Optimize}
\BlankLine
Initialize LVLM parameters $\boldsymbol{\theta}$ \;
Begin: $y_0=\texttt{BOS}, t=1$ \;
\BlankLine
\lWhile{$y_t \neq$ \texttt{EOS}}{\;
    \hskip1.0em $\hat{v} = \alpha * v_1 + (1 -\alpha) * v_2$ \tcp*{Modify image}
    \hskip1.0em $m_\theta = \frac{1}{\pi}\arccos{\left( \frac{\bar{f}_\theta(v) \cdot \bar{f}_\theta(\hat{v})}{\|\bar{f}_\theta(v)\|\|\bar{f}_\theta(\hat{v})\|} \right) }$ \tcp*{Sim margin}
    \hskip1.0em $\delta_\theta = P_\theta(y_t|v, x) + m_\theta  \ge P_\theta(y_t|\hat{v}, x) $ \;
    \hskip1.0em $P^\text{VORD}_\theta(y_t|v,\hat{v}, x) = \begin{cases}
        P_\theta(y_t|v, x), & \text{if } \delta_\theta \\
        0, & \text{otherwise}
    \end{cases}$ \\
    \hskip1.0em $P^\text{VORD}_\theta(y_t|v, \hat{v},  x) = 0, \text{if } y_t \notin \mathcal{V}(y_{<t})$ \tcp*{\cref{eq:adaptive_plausibility}}
    \BlankLine
    \Return $y_t \sim P^\text{VORD}_\theta(y_t|v,\hat{v},x, y_{<t})$
}
\BlankLine
\captionsetup{font=footnotesize}
\caption{VORD - Visual Ordinal Decoding}
\label{alg:vord_decoding_algorithm}
\end{algorithm}

\subsubsection{Adaptive Plausibility Constraints}
The ordinal relationship in \cref{eq:vord_decoding} may excessively penalize model outputs influenced by visually distorted inputs. Previous works have shown that although modified inputs tend to cause hallucinations, generated outputs often adhere to basic language rules and common sense \cite{VCD_2024_CVPR}. VORD may inadvertently penalize valid outputs if the modified image generates a token with higher confidence than the original. To address this issue, we adopt the adaptive plausibility constraints used by \cite{li-etal-2023-contrastive, VCD_2024_CVPR}:
\begin{equation}
    \begin{aligned}
    &\mathcal{V}(y_{<t}) = \{ y_t \in \mathcal{V}: \\
    &P_\theta(y_t|v, x, y_{<t}) \ge \beta \max_w P_\theta(w|v, x, y_{<t}) \}, \\
    &P^\text{VORD}_\theta(y_t|v, \hat{v},  x) = 0, \text{if } y_t \notin \mathcal{V}(y_{<t}),
    \label{eq:adaptive_plausibility}
    \end{aligned}
\end{equation}
where $\mathcal{V}$ represents the vocabulary of the LVLM and $\beta \in [0-1]$ is a hyperparameter controlling the strength of the truncation of the next token. Higher values of $\beta$ results more aggressive truncation, keeping only tokens with high confidence. The key steps for VORD decoding are summarized in \cref{alg:vord_decoding_algorithm}, which creates a modified image (\eg Mixup or Diffusion), followed by computing the similarity margin \cref{adaptive_sim_margin} and performs VORD and adaptive plausibility constraints respectively.

\subsection{Design 2: VORD Loss}
Following the transitive property introduced in \cref{eq:transitive}, we can derive a convex, piece-wise quadratic and differentiable loss function as follows:
\begin{equation}
    \mathcal{L}_{\text{vord}} = \max(P_\theta(y|\hat{v}, x) - P_\theta(y|v, x) + m_\theta, 0)^\psi
    \label{eq:vord_loss}
\end{equation}
where all tokens from the modified image with higher confidences than the original image, are penalized by a positive margin $m_\theta \ge 0$. The VORD loss can then be paired with the typical cross-entropy loss $\mathcal{L}$ during model optimization using gradient descent, ensuring that the model not only generates accurate text descriptions but also maintains consistent ordinal relationships between tokens. The final objective for VORD loss is thus given by: $\mathcal{L}_{\text{CE}} + \mathcal{L}_{\text{vord}}$. Note that VORD loss is only a transitive condition enforced on the model's probabilities and that the model does not directly learn on the modified images.

\subsubsection{Gradient Analysis}
We further discuss and analyze the VORD loss function, which can be broken down into the following two terms:
\begin{equation}
  \mathcal{L}_{\text{vord}} = 
  \begin{cases}
    (P_\theta(y|\hat{v}, x) - P_\theta(y|v, x) + m_\theta)^\psi, & \text{if }  g_\theta > 0 \\
    0, & \text{otherwise}
  \end{cases}
\end{equation}
where $\psi > 0$ is a power term and the function $g_\theta = P_\theta(y|\hat{v}, x) - P_\theta(y|v, x) + m_\theta$ is the positive penalty obtained by the model for violating the transitive property in VORD. In contrast, the model is not penalized if the transitive property is obeyed.

\noindent \textbf{First Derivative:} Using the chain rule, we obtain novel gradients of $\mathcal{L}_{\text{vord}}$ in the following form:
\begin{equation}
    \frac{\partial \mathcal{L}_{\text{vord}}}{\partial \theta} =
    \begin{cases}
    \psi g_\theta^{\psi-1} \frac{\partial g_\theta}{\partial \theta} & \text{if } g_\theta > 0 \\
    0, & \text{otherwise}
    \end{cases}
\end{equation}
which is simply the difference between the gradients of the clean and modified images, multiplied by the positive penalty function $g_\theta$. To further ensure that \cref{eq:vord_loss} is convex, we can compute the derivative of $\frac{\partial \mathcal{L}_{\text{vord}}}{\partial \theta}$.

\noindent \textbf{Second Derivative:} Using the product rule, the general form for the second derivative is obtained as:
\small
\begin{equation}
    \frac{ \partial^2 \mathcal{L}_{\text{vord}} }{\partial^2 \theta} =
    \begin{cases}
    \psi g_\theta^{\psi-1} \frac{\partial^2 g_\theta}{\partial^2 \theta} + (\psi^2 - \psi) g_\theta^{\psi-2} ( \frac{\partial g_\theta}{\partial \theta} )^2  & \text{if }  g_\theta > 0 \\
    0, & \text{otherwise}
    \end{cases}
\end{equation}
When $g_\theta >0$ the second derivative is always positive, ensuring convexity. When $g_\theta \le 0$ the second derivative is also zero, trivially indicating convexity. Therefore, the VORD loss in \cref{eq:vord_loss} is convex regardless of the value of $g_\theta$.

\subsection{Visual Similarities \& Adaptive Penalty Margin}
\label{adaptive_sim_margin}
How do we effectively determine the penalty margin? In cases where differences between the original and modified images are small, naively over-penalizing the model by a fixed margin can lead to poor model performance. Furthermore, the manual assignment of the margin hyperparameter typically requires a cumbersome grid-search in order to obtain the optimal values of $m_\theta$. 

As an alternative, we propose a novel adaptive penalty margin based on visual angular similarities between pairs of original and modified images. Consider the following function:
\begin{equation}
    m_\theta = \frac{1}{\pi}\arccos{\left( \frac{\bar{f}_\theta(v) \cdot \bar{f}_\theta(\hat{v})}{\|\bar{f}_\theta(v)\|\|\bar{f}_\theta(\hat{v})\|} \right)}
\end{equation}
where the cosine similarity is computed between the averaged original visual tokens $\bar{f}_\theta(v)$ and modified visual tokens $\bar{f}_\theta(\hat{v})$ extracted from the visual transformer $f_\theta$. The margin $m_\theta \in [0 - 1]$ is then given by the inverse cosine similarity, normalized by $\pi$ radians. This creates a general penalty margin that dynamically adjusts according to the level and type of noise in the modified image, ensuring appropriate penalization for any choice of image corruptions. Our proposed adaptive margin is designed to complement both VORD decoding and VORD loss, with \cref{fig:vord_overview} illustrating our entire workflow. 

\section{Experiments and Results}
\subsection{Datasets}
\paragraph{POPE} The Polling-based Object Probing Evaluation \cite{pope-2023-evaluating} benchmark queries the LVLM simple yes/no questions about whether certain specific objects exist in an input image. It uses a balanced set of questions (50-50 split of real and non-existent objects). There are three different settings used to determine non-existent objects: 1.) \textit{Random}, when non-existent objects are chosen randomly. 2.) \textit{Popular} where absent objects are selected from a list of commonly occurring objects and 3.) \textit{Adversarial} where co-occurring objects not in the image are selected. The POPE dataset is constructed from three datasets, namely: MSCOCO \cite{mscoco_Lin2014}, A-OKVQA \cite{A-OKVQA_Schwenk2022AOKVQAAB}, GQA \cite{gqa_Hudson_2019_CVPR}. We follow the evaluation protocols of other authors \cite{VCD_2024_CVPR} and report model performance using standard evaluation metics such as the accuracy, precision, recall, and F1 score.

\paragraph{MME Hallucination Subset} This benchmark comprehensively evaluates various LVLM capabilities, including ten perception tasks and four cognition tasks. It includes subsets specifically designed to assess object-level hallucination (existence and count) and attribute-level hallucination (position and color). The MME dataset is built upon six preceding works: MSCOCO \cite{mscoco_Lin2014}, MovieNet \cite{Huang2020MovieNetAH}, Places \cite{Places_neurips2014}, GoogleLandMarksV2 \cite{Weyand_2020_CVPR}, Art500K \cite{art500K_acm17} and CTW1500 \cite{ctw1500_LIU2019}. Model performances are measured using the accuracy and ``accuracy+" metrics.

\paragraph{LLaVA-Bench} This LLaVA benchmark \cite{LLaVA_neurips23, LLaVA_2024_CVPR} focuses on the LVLMs' ability to handle challenging tasks in various environments (indoor, outdoor, memes, paintings) and its ability to adapt to unseen domains. It includes 24 images comprising of 60 questions. We tailor specific case studies using the benchmark to qualitatively demonstrate the performances of different methods.

\begin{table*}[!tb]
\begin{center}
\begin{adjustbox}{width=0.8\textwidth}
\begin{tabular}{ccc|cccc|cc}
\hline
Datasets &Setting &Model &Method &Accuracy $\uparrow$ &Precision $\uparrow$ &Recall $\uparrow$ &F1 $\uparrow$ &ECE $\downarrow$\\
\hline
\multirow{36}{*}{A-OKVQA} & \multirow{12}{*}{Random} & \multirow{4}{*}{LLaVA-v1.5} & Regular &83.90$_{\pm0.29}$  &89.24$_{\pm0.41}$ &77.64$_{\pm0.36}$ &82.96$_{\pm0.30}$ &2.82$_{\pm0.11}$\\ 
& & & OPERA &84.35$_{\pm0.00}$ &89.72$_{\pm0.00}$ &78.09$_{\pm0.00}$ &83.43$_{\pm0.00}$ &2.94$_{\pm0.00}$\\
& & & VCD &86.17$_{\pm0.22}$ &90.76$_{\pm0.11}$ &81.07$_{\pm0.43}$ &85.54$_{\pm0.26}$ &7.35$_{\pm0.10}$\\
& & & VORD (Ours) &\cellcolor{blue!10}\textbf{88.35}$_{\pm0.32}$ &\cellcolor{blue!10}90.59$_{\pm0.33}$ &\cellcolor{blue!10}86.01$_{\pm0.30}$ &\cellcolor{blue!10} \textbf{88.18}$_{\pm0.32}$ &\cellcolor{blue!10}\textbf{2.69}$_{\pm0.09}$ \\
\cline{4-9}

 & & \multirow{4}{*}{Qwen-VL} & Regular &83.80$_{\pm0.16}$ &92.28$_{\pm0.25}$ &74.19$_{\pm0.10}$ &82.20$_{\pm0.16}$ &8.51$_{\pm0.29}$ \\
& & & OPERA &85.68$_{\pm0.00}$ &94.40$_{\pm0.00}$ &76.24$_{\pm0.00}$ &84.30$_{\pm0.00}$ &14.32$_{\pm0.00}$\\
& & & VCD &85.51$_{\pm0.09}$ &92.74$_{\pm0.01}$ &77.43$_{\pm0.23}$ &84.34$_{\pm0.13}$ &11.53$_{\pm0.07}$ \\
& & & VORD (Ours) &\cellcolor{blue!5}\textbf{85.87}$_{\pm0.13}$ &\cellcolor{blue!5}93.69$_{\pm0.28}$ &\cellcolor{blue!5}77.33$_{\pm0.21}$ &\cellcolor{blue!5}\textbf{84.67}$_{\pm0.14}$ &\cellcolor{blue!5}\textbf{7.56}$_{\pm0.14}$ \\
\cline{4-9}

 & & \multirow{4}{*}{InstructBLIP} & Regular &80.22$_{\pm0.31}$ &79.03$_{\pm0.32}$ &83.13$_{\pm0.63}$ &80.94$_{\pm0.33}$ &5.14$_{\pm0.28}$ \\
& & & OPERA &\textbf{85.07}$_{\pm0.00}$ &88.39$_{\pm0.00}$ &80.73$_{\pm0.00}$ &84.39$_{\pm0.00}$ &12.41$_{\pm0.00}$ \\
& & & VCD &83.30$_{\pm0.22}$  &82.93$_{\pm0.11}$ &84.70$_{\pm0.30}$  &83.66$_{\pm0.21}$  &6.69$_{\pm0.23}$ \\
& & & VORD (Ours) &\cellcolor{blue!10}84.19$_{\pm0.25}$ &\cellcolor{blue!10}84.37$_{\pm0.65}$ &\cellcolor{blue!10}84.70$_{\pm0.45}$ &\cellcolor{blue!10}\textbf{84.41}$_{\pm0.18}$ &\cellcolor{blue!10}\textbf{2.95}$_{\pm0.03}$\\
\cline{2-9}

 & \multirow{12}{*}{Popular} & \multirow{4}{*}{LLaVA-v1.5} &Regular &80.27$_{\pm0.52}$ &82.50$_{\pm0.56}$  &77.45$_{\pm0.56}$ &79.72$_{\pm0.54}$ &3.14$_{\pm0.04}$\\ 
& & & OPERA &81.04$_{\pm0.00}$ &83.36$_{\pm0.00}$ &78.16$_{\pm0.00}$ &80.50$_{\pm0.00}$ &3.23$_{\pm0.00}$\\
(Avg. scores of)& & & VCD &81.85$_{\pm0.37}$ &82.93$_{\pm0.32}$ &80.99$_{\pm0.45}$ &81.73$_{\pm0.38}$ &9.54$_{\pm0.03}$\\
MSCOCO & & & VORD (Ours) &\cellcolor{blue!5}\textbf{85.60}$_{\pm0.37}$ &\cellcolor{blue!5}85.90$_{\pm0.42}$ &\cellcolor{blue!5}85.53$_{\pm0.73}$ &\cellcolor{blue!5}\textbf{85.61}$_{\pm0.42}$ &\cellcolor{blue!5}2.83$_{\pm0.36}$\\
\cline{4-9}

+ & & \multirow{4}{*}{Qwen-VL} & Regular &81.70$_{\pm0.35}$ &87.53$_{\pm0.33}$ &74.45$_{\pm0.46}$ &80.34$_{\pm0.39}$ &9.09$_{\pm0.34}$ \\
& & & OPERA &83.02$_{\pm0.00}$ &90.61$_{\pm0.00}$ &76.24$_{\pm0.00}$ &82.71$_{\pm0.00}$ &15.98$_{\pm0.00}$\\
& & & VCD &83.09$_{\pm0.17}$ &87.6$_{\pm0.22}$ &77.67$_{\pm0.12}$ &82.22$_{\pm0.15}$ &13.12$_{\pm0.11}$\\
+ & & & VORD (Ours) &\cellcolor{blue!10}\textbf{83.96}$_{\pm0.14}$ &\cellcolor{blue!10}89.59$_{\pm0.23}$ &\cellcolor{blue!10}77.30$_{\pm0.15}$ &\cellcolor{blue!10}\textbf{82.86}$_{\pm0.15}$ &\cellcolor{blue!10}\textbf{8.25}$_{\pm0.23}$ \\
\cline{4-9}

GQA & & \multirow{4}{*}{InstructBLIP} & Regular &76.04$_{\pm0.31}$ &73.00$_{\pm0.50}$ &83.33$_{\pm0.24}$ &77.70$_{\pm0.20}$ &8.48$_{\pm0.33}$ \\
& & & OPERA &78.33$_{\pm0.00}$ &73.85$_{\pm0.00}$ &87.73$_{\pm0.00}$ &80.20$_{\pm0.00}$ &14.90$_{\pm0.00}$ \\
& & & VCD &78.86$_{\pm0.27}$ &76.13$_{\pm0.17}$ &84.85$_{\pm0.35}$ &80.09$_{\pm0.26}$ &10.24$_{\pm0.26}$ \\
& & & VORD (Ours) &\cellcolor{blue!5}\textbf{80.73}$_{\pm0.26}$ &\cellcolor{blue!5}78.76$_{\pm0.20}$ &8\cellcolor{blue!5}4.53$_{\pm0.31}$ &\cellcolor{blue!5}\textbf{81.43}$_{\pm0.26}$ &\cellcolor{blue!5} \textbf{3.74}$_{\pm0.13}$\\
\cline{2-9}

 & \multirow{12}{*}{Adversarial} & \multirow{4}{*}{LLaVA-v1.5} &Regular &76.39$_{\pm0.38}$ &76.36$_{\pm0.34}$ &77.42$_{\pm0.55}$ &76.68$_{\pm0.40}$  &5.63$_{\pm0.09}$\\ 
& & & OPERA &76.91$_{\pm0.00}$ &76.84$_{\pm0.00}$ &77.91$_{\pm0.00}$ &77.19$_{\pm0.00}$ &5.54$_{\pm0.00}$\\
& & & VCD &77.83$_{\pm0.03}$ &76.67$_{\pm0.15}$ &81.17$_{\pm0.33}$ &78.61$_{\pm0.09}$ &12.39$_{\pm0.04}$\\
& & & VORD (Ours) &\cellcolor{blue!10}\textbf{80.72}$_{\pm0.71}$ &\cellcolor{blue!10}78.51$_{\pm0.63}$ &\cellcolor{blue!10}85.27$_{\pm1.02}$ &\cellcolor{blue!10}\textbf{81.60}$_{\pm0.73}$ &\cellcolor{blue!10}\textbf{4.57}$_{\pm0.30}$\\
\cline{4-9}

 & & \multirow{4}{*}{Qwen-VL} & Regular &79.46$_{\pm0.06}$ &82.94$_{\pm0.10}$ &74.73$_{\pm0.13}$ &78.47$_{\pm0.08}$ &10.86$_{\pm0.13}$ \\
& & & OPERA &80.43$_{\pm0.00}$ &85.48$_{\pm0.00}$ &76.24$_{\pm0.00}$ &80.43$_{\pm0.00}$ &18.57$_{\pm0.00}$\\
& & & VCD &80.79$_{\pm0.12}$ &83.32$_{\pm0.05}$ &77.45$_{\pm0.28}$ &80.15$_{\pm0.17}$ &15.21$_{\pm0.13}$ \\
& & & VORD (Ours) &\cellcolor{blue!5}\textbf{81.22}$_{\pm0.11}$ &\cellcolor{blue!5}84.22$_{\pm0.22}$ &\cellcolor{blue!5}77.45$_{\pm0.16}$ &\cellcolor{blue!5}\textbf{80.52}$_{\pm0.07}$ &\cellcolor{blue!5}\textbf{10.09}$_{\pm0.06}$ \\
\cline{4-9}

 & & \multirow{4}{*}{InstructBLIP} & Regular &72.71$_{\pm0.27}$ &69.21$_{\pm0.29}$  &82.99$_{\pm0.25}$  &75.31$_{\pm0.21}$  &11.11$_{\pm0.27}$ \\
& & & OPERA &75.50$_{\pm0.00}$ &70.49$_{\pm0.00}$ &87.73$_{\pm0.00}$ &78.17$_{\pm0.00}$ &17.53$_{\pm0.00}$\\
& & & VCD &74.81$_{\pm0.21}$ &71.27$_{\pm0.22}$ &84.67$_{\pm0.35}$ &77.16$_{\pm0.20}$ &13.38$_{\pm0.18}$ \\
& & & VORD (Ours) &\cellcolor{blue!10}\textbf{76.77}$_{\pm0.35}$ &\cellcolor{blue!10}73.67$_{\pm0.41}$ &\cellcolor{blue!10}84.51$_{\pm0.17}$ &\cellcolor{blue!10}\textbf{78.51}$_{\pm0.30}$ &\cellcolor{blue!10}\textbf{6.82}$_{\pm0.05}$\\
\hline
\end{tabular}
\end{adjustbox}
\end{center}
\vspace{-1.0em}
\caption{We report the averaged results (\%) of reruns for different methods evaluated on all three datasets of the POPE benchmark. VORD decoding effectively mitigates object hallucinations, while better calibrating model predictions.}
\label{table:pope_results}
\end{table*}

\subsection{Models \& Experiment Setup}
We compare VORD Decoding against recently published methods, including VCD \cite{VCD_2024_CVPR}, OPERA beam search \cite{OPERA_2024_CVPR} and a baseline using regular decoding. For our evaluation on image question-answering benchmarks, we use LLaVA-v1.5 \cite{LLaVA_2024_CVPR}, Qwen-VL and InstructBLIP \cite{instructblip_dai2023} backbone LVLMs. For VORD loss, we perform instruction finetuning of LLaVA-v1.5 (7B and 13B) \cite{LLaVA_2024_CVPR} initialized from pretrained Vicuna-v1.5 weights \cite{LLAMA_2023} and keep all hyperparameters on default as per detailed by \cite{LLaVA_2024_CVPR}. We fix the number of bins as B=15 for the ECE. Our decoding experiments are evaluated on two NVIDIA A100 GPUs, while models trained with VORD loss are conducted on two NVIDIA H100 GPUs. More details on hyperparameters are included in \cref{appendix:setup}.

\subsection{Experiment Results}
\paragraph{Performance gains with VORD Loss}
We evaluate the effectiveness of our method by applying the transitive properties of VORD loss to fine-tune LoRA-adapted \cite{hu2022lora} backbone LVLM architectures; LLaVA-7B and LLaVA-13B \cite{LLaVA_neurips23} using our proposed VORD loss. Specifically, we compare the performance of a baseline model trained only with cross-entropy loss and models trained with VORD loss using different power terms ($\psi=1$ and $\psi=2$). Additionally, we show the performances of these models, when VORD decoding is further applied during inference.

\begin{figure}[!tb]
    \centering
    \includegraphics[width=\columnwidth]{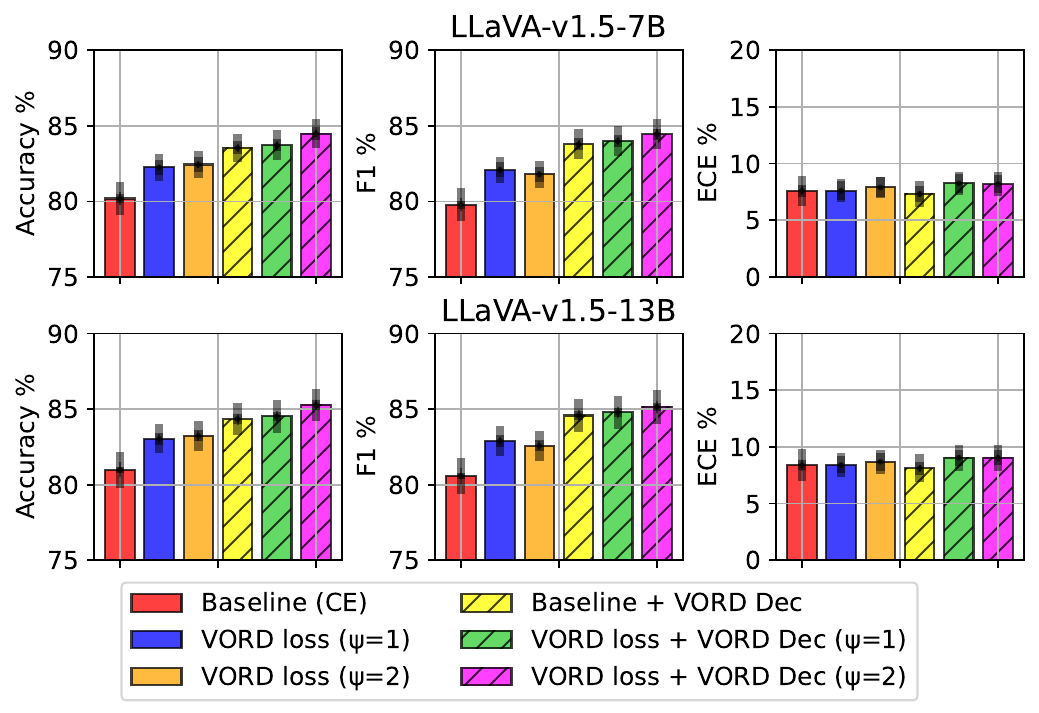}
    \vspace{-2.0em}
    \caption{Our experiments demonstrate that finetuning LLaVA with VORD loss, particularly the squared variant $(\psi=2)$, highlighted in magenta, leads to consistent performance gains over the baseline. Moreover, combining VORD loss with VORD decoding (shaded) results in additional improvements.}
    \label{fig:vord_loss_results}
\end{figure}

Our results in \cref{fig:vord_loss_results} demonstrate that VORD loss yields significant improvements, particularly the squared variant ($\psi=2$), by boosting accuracy and F1-score on the POPE hallucination benchmark by up to +2.9\% and +2.7\%, respectively, while maintaining good ECE throughout.  Furthermore, combining VORD loss with VORD decoding results in additional gains, achieving the best overall performance. These improvements are consistent across both LLaVA-v1.5-7B and LLaVA-v1.5-13B. Additional results and discussions for VORD loss are presented in \cref{appendix:additional_results}.

\paragraph{Performance gains with VORD Decoding}
In \cref{table:pope_results}, we compare our proposed VORD decoding against recently published baselines evaluated on the POPE benchmark. Our experiments demonstrate that VORD decoding achieves the state-of-the-art performance on the POPE benchmark, highlighting the effectiveness of VORD in mitigating object hallucinations. Specifically, we report consistent improvements over baseline methods by margins of up to +2.2\% in accuracy and +2.3\% in F1. Additionally, VORD exhibits better calibration, as evidenced by the best ECE scores on LLaVA, QWEN and InstructBLIP. 

Furthermore, all LVLMs exhibit a significant performance decline when transitioning from random to popular and adversarial settings. In contrast, VORD remains relatively robust, even in the highly challenging adversarial subset. Additionally, \cref{table:mme_results} presents a comparative analysis of various methods on the MME perception benchmark. While improvements vary across individual tasks, we observe significant gains over regular decoding in overall performance for LLaVA, QWEN, and InstructBLIP of 7.6\%, 5.5\%, and 15.9\% respectively. Notably, VORD achieves the highest results overall securing the top performance and calibration on all architectures. Individual scores of each method are detailed in \cref{appendix:additional_results}.

\begin{table}[!tb]
\large
\begin{center}
\begin{adjustbox}{width=1.0\columnwidth}
\begin{tabular}{c|ccccc}
\toprule
Model &Method &Perception$\uparrow$ &Recognition$\uparrow$ &Overall Scores$\uparrow$\\
\midrule
\multirow{4}{*}{LLaVA-v1.5}
&Regular &1294.86$_{\pm21.20}$ &331.19$_{\pm15.05}$ &1626.05$_{\pm28.82}$ \\
&VCD &1356.27$_{\pm20.05}$ &309.04$_{\pm21.89}$ &1665.32$_{\pm33.45}$  \\
&OPERA &1332.79$_{\pm0.00}$ &330.71$_{\pm0.00}$ &1663.50$_{\pm0.00}$\\
&VORD (Ours) &\cellcolor{blue!5}\textbf{1393.75}$_{\pm20.29}$ &\cellcolor{blue!5}334.04$_{\pm20.77}$ &\cellcolor{blue!5}\textbf{1727.79}$_{\pm32.47}$\\
\midrule
\multirow{4}{*}{Qwen-VL}
&Regular &1352.50$_{\pm21.31}$ &313.04$_{\pm16.40}$ &1665.54$_{\pm36.62}$ \\
&VCD &1403.17$_{\pm14.57}$ &318.13$_{\pm22.77}$ &1721.30$_{\pm36.16}$ \\
&OPERA &1400.37$_{\pm25.86}$ &271.78$_{\pm31.20}$ &1672.15$_{\pm42.66}$ \\
&VORD (Ours) &\cellcolor{blue!10}\textbf{1427.07}$_{\pm22.90}$ &\cellcolor{blue!10}314.28$_{\pm21.75}$ &\cellcolor{blue!10}\textbf{1741.36}$_{\pm36.73}$\\
\midrule
\multirow{4}{*}{InstructBLIP}
&Regular &976.99$_{\pm27.70}$ &276.42$_{\pm17.10}$ &1253.42$_{\pm28.26}$  \\
&VCD &1088.80$_{\pm28.58}$ &292.50$_{\pm18.16}$ &1381.30$_{\pm30.64}$ \\
&OPERA &1044.80$_{\pm0.00}$ &270.35$_{\pm0.00}$ &1315.15$_{\pm0.00}$\\
&VORD (Ours) &\cellcolor{blue!5}\textbf{1132.98}$_{\pm29.06}$ &\cellcolor{blue!5}271.54$_{\pm12.93}$ &\cellcolor{blue!5}\textbf{1404.52}$_{\pm32.74}$ \\
\midrule
\end{tabular}
\end{adjustbox}
\end{center}
\vspace{-1.0em}
\caption{We report the main results (\%) comparing different methods computed across 3 seeds, evaluated on the MME benchmark. VORD decoding delivers the best overall performance.}  
\label{table:mme_results}
\end{table}

\begin{figure*}[!tb]
    \centering
    \includegraphics[width=\textwidth]{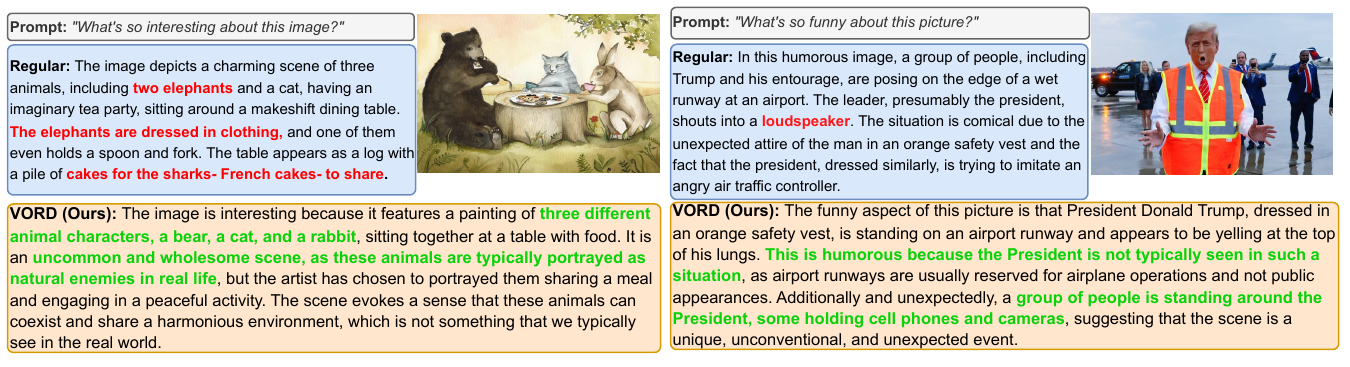}
    \vspace{-2.0em}
    \caption{VORD produces accurate and detailed outputs, mitigating object hallucinations across a wide range of different visual cues. Hallucinations are in \textcolor{red}{red}, while accurate and detailed responses are in \textcolor{green}{green}.}
    \label{fig:vord_examples}
\end{figure*}

\section{Discussion and Analysis}
\paragraph{Evaluation and Analysis on LLaVA-Bench}
We further examine VORD's performance on complex open-ended VQA tasks, using LLaVA-Bench. We adopt the evaluation protocol of \citep{Yin2023WoodpeckerHC, VCD_2024_CVPR} and select the latest LVLM model GPT4oV as a third party evaluator. For our assessment, we specifically prompt each backbone LVLM with ``\textit{Describe this image in detail.}" for every image. We then consolidate the responses of each decoding method and score individual responses on GPT4oV using a prompt card (see \cref{appendix:case_studies}). GPT4oV is tasked to rate the accuracy and detailedness of responses, paying extra attention to any hallucinations in the generations. Accuracy refers to the correctness of the response, while Detailedness measures how informative each response is. \cref{table:llava_bench_results} shows that VORD consistently produces highly accurate and detailed responses across all three baseline models, demonstrating its effectiveness in tackling complex VQA tasks.
\begin{table}[!tb]
\centering
\adjustbox{width=0.9\columnwidth}{
\begin{tabular}{c|cc c c}
\toprule
Model &Method &Avg Lengths &Accuracy$\uparrow$ &Detailedness$\uparrow$\\
\midrule
\multirow{3}{*}{LLaVA-1.5} & Regular &97.02$_{\pm2.78}$ &4.41 &4.75 \\
& VCD &97.93$_{\pm3.98}$ &5.20 &5.50 \\
& VORD &\cellcolor{blue!5}91.61$_{\pm6.32}$ &\cellcolor{blue!5}\textbf{5.88} &\cellcolor{blue!5}\textbf{5.92}\\
\midrule
\multirow{3}{*}{InstructBLIP} &Regular &102.21$_{\pm1.20}$ &3.39 &3.83  \\
& VCD &104.00$_{\pm1.04}$ &4.17 &4.69 \\
& VORD &\cellcolor{blue!10}97.91$_{\pm1.15}$ &\cellcolor{blue!10}\textbf{4.65} &\cellcolor{blue!10}\textbf{4.70}\\
\midrule
\multirow{3}{*}{Qwen-VL} & Regular &10.41$_{\pm1.21}$ &7.28 &6.09 \\
& VCD &10.36$_{\pm1.15}$ &7.30 &6.76 \\
& VORD &\cellcolor{blue!5}13.13$_{\pm1.62}$ &\cellcolor{blue!5}\textbf{8.86} &\cellcolor{blue!5}\textbf{6.95}\\
\bottomrule
\end{tabular}
}
\caption{Evaluation results from GPT-4oV comparing the open-ended responses for different methods. Accuracy and Detailedness are gauged on a scale of 10. VORD delivers short but accurate and detailed responses.}
\label{table:llava_bench_results}
\end{table}
\paragraph{Length of token generations (Short \& Sweet)}
A concept known as Occam's Razor states that when you have two choices that are equally good, the simpler one is usually better. Building on this principle, we explore how VORD generates shorter text outputs by comparing the average token generation lengths of regular decoding, VCD, and VORD on LLaVA-Bench.  \cref{table:llava_bench_results} shows that VORD has average shorter token generation lengths, whilst achieving the highest accuracy and detailedness. 

Delving deeper, we analyze two case studies in \cref{fig:vord_examples}. In the first case, regular decoding hallucinates potentially present objects like $\texttt{<elephants>}$ and $\texttt{<cakes>}$, whereas VORD accurately identifies the three distinct animals in the image. In the second case involving an image of Trump, regular decoding hallucinates the object $\texttt{<loudspeaker>}$, whereas VORD accurately captures the uniqueness of President Trump's antics. 
\begin{table}[!tb]
\begin{center}
\begin{adjustbox}{width=1.0\columnwidth}
\begin{tabular}{cc|cccc}
\hline
&$m_\theta$ &Perception$\uparrow$ &Recognition$\uparrow$ &Overall Scores$\uparrow$\\
\midrule
&Regular &1294.86$_{\pm21.20}$ &331.19$_{\pm15.05}$ &1626.05$_{\pm28.82}$ \\
&0.00 &1172.51$_{\pm15.54}$ &245.71$_{\pm21.55}$ &1418.23$_{\pm30.68}$\\
&0.25 &1376.61$_{\pm20.87}$ &342.85$_{\pm18.68}$ &1719.46$_{\pm31.09}$ \\
&0.50 &1356.23$_{\pm22.72}$ &351.25$_{\pm16.80}$ &1707.48$_{\pm30.26}$\\
&0.75 &1336.50$_{\pm23.50}$ &350.89$_{\pm16.67}$ &1687.39$_{\pm30.07}$ \\
&Adaptive (Ours) &\cellcolor{blue!5}\textbf{1393.75}$_{\pm20.29}$ &\cellcolor{blue!5}334.04$_{\pm20.77}$ &\cellcolor{blue!5}\textbf{1727.79}$_{\pm32.47}$\\
\midrule
\end{tabular}
\end{adjustbox}
\end{center}
\vspace{-1.0em}
\caption{Comparisons of different margin settings for VORD decoding evaluated on MME. Our proposed adaptive visual similarity margin delivers the best performance.}
\label{table:mme_margin_ablations}
\end{table}
\paragraph{Effects of Margin Parameter}
We perform an ablation study to evaluate the influence of our proposed adaptive visual similarity margin on VORD decoding. Specifically, we compare our method against regular decoding and four fixed margin settings: $m_\theta = [0.00, 0.25, 0.50, 0.75]$ using LLaVA-v1.5 on the MME benchmark. \cref{table:mme_margin_ablations} shows that VORD decoding without a margin $(m_\theta=0.0)$, struggles to differentiate tokens between original and modified images, leading to weaker performance than regular decoding. Conversely, excessively large margins $(m_\theta>0.25)$ can over-penalize predictions, degrading performance. Our proposed adaptive margin, which eliminates the need for hyperparameter tuning, achieves the best overall performance. Additional ablation studies on the effects of various visual corruptions are included in \cref{appendix:additional_ablation_studies}.

\section{Limitations}
While this work primarily focuses on mitigating object hallucinations in LVLMs, we believe that the transitive property of VORD can be extended to LLMs and NLP. Although both VORD decoding and VORD loss currently rely on image modifications, we envision future adaptations involving text or ground truth label modifications. This would enable VORD to address hallucinations in a broader range of applications.

\section{Conclusion}
We present VORD, visual ordinal calibration for LVLMs. VORD is proposed in two forms: a training-free contrastive decoding method and a novel ordinal ranking loss function. VORD also utilizes an adaptive visual similarity margin, which dynamically computes a penalty margin based on the visual similarities between original and modified images. Our experiments demonstrate VORD's effectiveness in mitigating object hallucinations and delivering well-calibrated confidence estimates in LVLMs.
{
    \small
    \bibliographystyle{ieeenat_fullname}
    \bibliography{main}
}
\newpage
\appendix
\section{Supplementary Experiments and Results}
\label{appendix:additional_results}

\subsection{Experiment Details}
\label{appendix:setup}
\paragraph{Hyperparameter settings} In general, we follow the default hyperparameters of each algorithm as proposed by the original authors. \cref{table:hyperparameters} lists the key hyperparameters used in our experiments for both VORD decoding and VORD loss.

\begin{table}[htb]
\center
\begin{adjustbox}{width=0.6\textwidth}
\begin{tabular}{cc|c}
&\textbf{Hyperparameters} &\textbf{Values} \\
\hline
&Learning rate &1e-5 \\
&Learning rate scheduler &Cosine \\
&Batch size per GPU &8 \\
&Optimizer &Adam \\
&Epochs &1\\
&Temperature (T) &1.0\\
&$\beta$ &0.2 \\
&$\psi$ &1.0 or 2.0 \\
&Top P &1.0 \\
&Top K &None \\
&Mixup ($\alpha$) &1.0 \\
&No. of bins $B$ &15.0\\
\end{tabular}
\end{adjustbox}
\caption{Hyperparameters used in our experiments for VORD.}
\label{table:hyperparameters}
\end{table}

\begin{table*}[!htb]
\centering
\begin{adjustbox}{width=0.95\textwidth}
\begin{tabular}{ccccccccccccc}
\hline
\multirow{2}{*}{Model}
&\multirow{2}{*}{Method} &\multicolumn{10}{c}{MME Tasks (Perception)} &\multirow{2}{*}{Overall Scores$\uparrow$}\\
& &Existence &Count &Position &Color &Posters &Celebrity &Scene &Landmark &Artwork &OCR \\
\hline
\multirow{4}{*}{LLaVA-v1.5}
&Regular &181.67$_{\pm2.36}$  &111.67$_{\pm19.77}$ &118.89$_{\pm1.57}$ &140.56$_{\pm6.98}$ &124.60$_{\pm3.36}$ &112.65$_{\pm1.44}$ &147.58$_{\pm3.91}$ &131.25$_{\pm3.89}$  &108.50$_{\pm4.64}$ &117.50$_{\pm10.8}$  &1294.86$_{\pm21.20}$ \\
&VCD &181.67$_{\pm2.36}$ &122.78$_{\pm13.63}$ &121.67$_{\pm1.36}$ &143.89$_{\pm6.29}$ &126.98$_{\pm3.85}$ &134.71$_{\pm2.60}$ &155.58$_{\pm1.65}$ &141.58$_{\pm0.51}$ &112.42$_{\pm2.26}$ &115.00$_{\pm3.54}$  &1356.27$_{\pm20.05}$\\
&OPERA &180.67$_{\pm0.00}$ &133.33$_{\pm0.00}$ &123.33$_{\pm0.00}$ &155.00$_{\pm0.00}$ &134.69$_{\pm0.00}$ &116.76$_{\pm0.00}$ &152.75$_{\pm0.00}$ &133.01$_{\pm0.00}$ &103.25$_{\pm0.00}$ &100.00$_{\pm0.00}$ &1332.79$_{\pm0.00}$\\
&VORD (Ours) &\cellcolor{blue!5}185.00$_{\pm4.08}$ &\cellcolor{blue!5}132.22$_{\pm11.57}$ &\cellcolor{blue!5}117.22$_{\pm3.42}$ &\cellcolor{blue!5}145.56$_{\pm9.06}$ &\cellcolor{blue!5}133.45$_{\pm3.93}$ &\cellcolor{blue!5}141.47$_{\pm3.39}$ &\cellcolor{blue!5}154.17$_{\pm1.03}$ &\cellcolor{blue!5}151.67$_{\pm1.01}$ &\cellcolor{blue!5}117.17$_{\pm4.85}$ &\cellcolor{blue!5}115.83$_{\pm6.29}$ &\cellcolor{blue!5}\textbf{1393.75}$_{\pm20.29}$\\
\hline
\multirow{4}{*}{Qwen-VL}
&Regular &145.00$_{\pm15.0}$ &115.83$_{\pm2.50}$ &122.5$_{\pm5.83}$ &176.67$_{\pm1.67}$ &136.73$_{\pm6.46}$ &121.76$_{\pm0.29}$ &148.62$_{\pm1.62}$ &159.12$_{\pm2.62}$ &125.00$_{\pm2.00}$ &101.25$_{\pm1.25}$ &1352.49$_{\pm21.31}$\\
&VCD  &156.00$_{\pm6.52}$ &131.00$_{\pm6.19}$ &128.00$_{\pm3.61}$ &181.67$_{\pm5.14}$ &142.45$_{\pm2.96}$ &137.35$_{\pm2.45}$ &149.10$_{\pm2.51}$ &163.95$_{\pm1.77}$ &127.65$_{\pm2.81}$ &86.00$_{\pm3.35}$ &1403.17$_{\pm14.57}$ \\
&OPERA  &165.00$_{\pm0.00}$ &118.33$_{\pm0.00}$ &138.33$_{\pm0.00}$ &180.00$_{\pm0.00}$ &142.18$_{\pm0.00}$ &118.53$_{\pm0.00}$ &157.00$_{\pm0.00}$ &160.75$_{\pm0.00}$ &132.75$_{\pm0.00}$ &87.50$_{\pm0.00}$ &1400.37$_{\pm0.00}$ \\
&VORD (Ours) &\cellcolor{blue!10}165.00$_{\pm0.10}$ &\cellcolor{blue!10}120.56$_{\pm5.67}$ &\cellcolor{blue!10}128.33$_{\pm8.16}$ &\cellcolor{blue!10}181.67$_{\pm2.36}$ &\cellcolor{blue!10}143.65$_{\pm3.53}$ &\cellcolor{blue!10}129.71$_{\pm2.53}$ &\cellcolor{blue!10}154.42$_{\pm1.53}$ &\cellcolor{blue!10}166.17$_{\pm3.16}$ &\cellcolor{blue!10}135.08$_{\pm3.14}$ &\cellcolor{blue!10}102.50$_{\pm6.12}$ &\cellcolor{blue!10}\textbf{1427.07}$_{\pm22.90}$\\
\hline
\multirow{4}{*}{InstructBLIP}
&Regular &153.33$_{\pm0.10}$ &80.00$_{\pm0.10}$ &65.00$_{\pm0.10}$ &108.33$_{\pm0.10}$ &98.98$_{\pm0.10}$ &97.35$_{\pm2.16}$ &130.50$_{\pm0.10}$ &103.25$_{\pm0.10}$ &85.25$_{\pm5.59}$ &55.00$_{\pm0.10}$ &976.99$_{\pm27.70}$ \\
&VCD &162.78$_{\pm6.98}$ &90.00$_{\pm4.08}$ &65.56$_{\pm4.16}$ &118.89$_{\pm10.30}$ &109.30$_{\pm2.80}$ &120.78$_{\pm3.96}$ &134.33$_{\pm2.70}$ &125.83$_{\pm3.68}$ &92.17$_{\pm4.12}$ &69.17$_{\pm11.61}$ &1088.80$_{\pm28.58}$\\
&OPERA  &175.00$_{\pm0.00}$ &55.00$_{\pm0.00}$ &50.00$_{\pm0.00}$ &118.15$_{\pm0.00}$ &122.86$_{\pm0.00}$ &80.00$_{\pm0.00}$ &149.25$_{\pm0.00}$ &138.79$_{\pm0.00}$ &90.75$_{\pm0.00}$ &65.00$_{\pm0.00}$ &1044.80$_{\pm0.00}$\\
&VORD (Ours) &\cellcolor{blue!5}164.44$_{\pm4.78}$  &\cellcolor{blue!5}85.00$_{\pm1.36}$ &\cellcolor{blue!5}65.00$_{\pm1.36}$ &\cellcolor{blue!5}134.44$_{\pm3.42}$ &\cellcolor{blue!5}113.95$_{\pm5.42}$ &\cellcolor{blue!5}119.31$_{\pm2.93}$ &\cellcolor{blue!5}138.67$_{\pm5.42}$ &\cellcolor{blue!5}132.75$_{\pm5.62}$ &\cellcolor{blue!5}96.08$_{\pm2.97}$ &\cellcolor{blue!5}83.33$_{\pm4.25}$ &\cellcolor{blue!5}\textbf{1132.98}$_{\pm29.06}$ \\
\hline

\end{tabular}
\end{adjustbox}
\caption{Detailed results (\%) comparing different decoding methods computed across 3 seeds, evaluated on the perception subset of MME.}  
\label{table:full_mme_results}
\end{table*}

\subsection{VORD Decoding}
We detail the individual scores of VORD decoding evaluated on MME (Perception) tasks. \cref{table:full_mme_results} shows the full split of all ten individual perception tasks of MME, evaluated on all three backbone architectures. Our findings show that solely using regular decoding results in the lowest scores, with improvements obtained when using VCD and OPERA, with  VORD decoding achieving the highest overall scores. 

An interesting observation is that the scores of individual tasks tend to vary across algorithms. For example, on the LLaVA-v-1.5 backbone, we can see that VORD decoding performs best on the Existence task with 185 points, while achieving the lowest score on Position with only 117.22 points. Ideally, we would expect a general improvement across all tasks, leading to the highest overall scores. A possible explanation to this variance in results could be due to the randomness of algorithms used, such Mixup or Diffusion during image modifications and multinomial sampling during the generation process.

\subsection{VORD Loss}
We further analyze the improvements of VORD loss evaluated on the POPE and MME benchmarks. \cref{table:vord_loss_pope_results} shows the averaged scores of all three datasets of POPE across the random, popular and adversarial settings. We compare the scores of two model backbones LLaVA-v1.5-7B and LLaVA-v1.5-13B, finetuned with and without VORD loss using regular decoding. 

Our results consistently demonstrate clear improvements when using VORD loss, with performance gains of roughly +2.5\% and +2.0\% in model accuracy and F1. While both $\psi=1$ and $\psi=2$ yield similar improvements, the squared variant $\psi=2$ exhibits a slight edge. Based on these findings, we therefore recommend using the squared variant of VORD loss for optimal performance. For completeness, the algorithm of VORD loss for next-word prediction is presented in \cref{alg:vord_loss_algorithm}.

\begin{algorithm}[!tb]
\SetAlgoNoLine
\SetAlgoNoEnd
\DontPrintSemicolon
\KwData{Given training corpus $D_\text{train} = (x_i, y_i)^N_{i=1}$ }
\SetKwFunction{fOptimize}{Optimize}
\BlankLine
Initialize LVLM parameters $\boldsymbol{\theta}$ \;
Begin: $y_0=\texttt{BOS}, t=1$ \;
\BlankLine
\lFor{$e \in epochs$}{\;
    \hskip1.0em $\hat{v} = \alpha * v_1 + (1 -\alpha) * v_2$ \tcp*{Modify image}
    \hskip1.0em $m_\theta = \frac{1}{\pi}\arccos{\left( \frac{\bar{f}_\theta(v) \cdot \bar{f}_\theta(\hat{v})}{\|\bar{f}_\theta(v)\|\|\bar{f}_\theta(\hat{v})\|} \right) }$ \tcp*{Sim margin}
    \BlankLine
    \hskip1.0em $P_\theta(y|v, x) = \boldsymbol{\theta}(v, x)$ \tcp*{Clean image}
    \hskip1.0em $P_\theta(y|\hat{v}, x) = \boldsymbol{\theta}(\hat{v}, x)$ \tcp*{Modified image}
    \hskip1.0em $\mathcal{L}_{\text{vord}} = \max(P_\theta(y|\hat{v}, x) - P_\theta(y|v, x) + m_\theta, 0)^\psi$\;
    \hskip1.0em $\mathcal{L}_{\text{total}} = \mathcal{L}_{\text{CE}} + \mathcal{L}_{\text{VORD}}$ \;
    \hskip1.0em $\boldsymbol{\theta_\text{new}} \leftarrow \boldsymbol{\theta_\text{old}} - \eta \nabla_{\boldsymbol{\theta}} \mathcal{L}_{\text{total}} $ \tcp*{Update parameters $\boldsymbol{\theta}$}
    \Return $\boldsymbol{\theta}$
}
\BlankLine
\captionsetup{font=footnotesize}
\caption{VORD loss}
\label{alg:vord_loss_algorithm}
\end{algorithm}

\begin{table*}[!ht]
\center
\begin{adjustbox}{width=0.85\textwidth}
\resizebox{\textwidth}{!}{
\begin{tabular}{ccccccc|cc}
\hline
Datasets &Setting &Model &Method &Accuracy $\uparrow$ &Precision $\uparrow$ &Recall $\uparrow$ &F1 $\uparrow$ &ECE $\downarrow$\\
\hline
\multirow{25}{*}{A-OKVQA} & \multirow{6}{*}{Random} & \multirow{3}{*}{LLaVA-v1.5-7B} & Baseline &83.90$_{\pm0.29}$  &89.24$_{\pm0.41}$ &77.64$_{\pm0.36}$ &82.96$_{\pm0.30}$ &2.82$_{\pm0.11}$\\
& & & VORD loss ($\psi=1$) &\cellcolor{blue!5}\textbf{86.46}$_{\pm0.12}$ &\cellcolor{blue!5}91.62$_{\pm0.27}$ &\cellcolor{blue!5}80.81$_{\pm0.20}$ &\cellcolor{blue!5} \textbf{85.74}$_{\pm0.12}$ &\cellcolor{blue!5}2.94$_{\pm0.15}$ \\ 
& & & VORD loss ($\psi=2$) &\cellcolor{blue!10}85.80$_{\pm0.19}$ &\cellcolor{blue!10}92.41$_{\pm0.36}$ &\cellcolor{blue!10}78.54$_{\pm0.17}$ &\cellcolor{blue!10}84.78$_{\pm0.18}$ &\cellcolor{blue!10}3.01$_{\pm0.12}$ \\ 
\cline{4-9}
 
 & & \multirow{3}{*}{LLaVA-v1.5-13B} &Baseline &84.20$_{\pm0.23}$ &89.08$_{\pm0.46}$ &78.46$_{\pm0.28}$ &\textbf{83.36}$_{\pm0.22}$ &2.65$_{\pm0.08}$ \\
& & & VORD loss ($\psi=1$) &\cellcolor{blue!5}85.83$_{\pm0.26}$ &\cellcolor{blue!5}91.63$_{\pm0.29}$ &\cellcolor{blue!5}79.41$_{\pm0.49}$ &\cellcolor{blue!5}84.95$_{\pm0.30}$ &\cellcolor{blue!5}2.74$_{\pm0.08}$\\ 
& & & VORD loss ($\psi=2$) &\cellcolor{blue!10}\textbf{85.85}$_{\pm0.12}$ &\cellcolor{blue!10}92.05$_{\pm0.13}$ &\cellcolor{blue!10}79.00$_{\pm0.36}$ &\cellcolor{blue!10} \textbf{85.03}$_{\pm0.15}$ &\cellcolor{blue!10}3.04$_{\pm0.08}$ \\
\cline{2-9}

 & \multirow{6}{*}{Popular} & \multirow{3}{*}{LLaVA-v1.5-7B} &Baseline &80.27$_{\pm0.52}$ &82.50$_{\pm0.56}$  &77.45$_{\pm0.56}$ &79.72$_{\pm0.54}$ &3.14$_{\pm0.04}$\\
(Avg. scores of)  & & & VORD loss ($\psi=1$) &\cellcolor{blue!5}82.06$_{\pm0.24}$ &\cellcolor{blue!5}83.82$_{\pm0.25}$ &\cellcolor{blue!5}80.4$_{\pm0.27}$ &\cellcolor{blue!5}81.79$_{\pm0.26}$ &\cellcolor{blue!5}3.31$_{\pm0.06}$ \\  
MSCOCO & & & VORD loss ($\psi=2$) &\cellcolor{blue!10}\textbf{82.78}$_{\pm0.32}$  &\cellcolor{blue!10}86.38$_{\pm0.43}$  &\cellcolor{blue!10}78.47$_{\pm0.24}$  &\cellcolor{blue!10}\textbf{81.99}$_{\pm0.31}$ &\cellcolor{blue!10}3.56$_{\pm0.06}$ \\  
\cline{4-9}

+ & & \multirow{3}{*}{LLaVA-v1.5-13B} &Baseline &81.83$_{\pm0.40}$ &84.48$_{\pm0.60}$ &78.34$_{\pm0.41}$ &81.16$_{\pm0.39}$ &2.85$_{\pm0.08}$ \\
GQA & & & VORD loss ($\psi=1$) &\cellcolor{blue!5}81.53$_{\pm0.12}$ &\cellcolor{blue!5}83.93$_{\pm0.17}$ &\cellcolor{blue!5}78.91$_{\pm0.07}$ &\cellcolor{blue!5}81.07$_{\pm0.10}$ &\cellcolor{blue!5}2.96$_{\pm0.08}$\\ 
+ & & & VORD loss ($\psi=2$) &\cellcolor{blue!10}\textbf{82.05}$_{\pm0.12}$ &\cellcolor{blue!10}84.65$_{\pm0.13}$ &\cellcolor{blue!10}79.12$_{\pm0.36}$ &\cellcolor{blue!10} \textbf{81.79}$_{\pm0.15}$ &\cellcolor{blue!10}2.94$_{\pm0.08}$ \\
\cline{2-9}

 & \multirow{6}{*}{Adversarial} & \multirow{3}{*}{LLaVA-v1.5-7B} &Baseline &76.39$_{\pm0.38}$ &76.36$_{\pm0.34}$ &77.42$_{\pm0.55}$ &76.68$_{\pm0.40}$ &5.63$_{\pm0.09}$\\
& & & VORD loss ($\psi=1$) &\cellcolor{blue!5}78.22$_{\pm0.21}$ &\cellcolor{blue!5}77.91$_{\pm0.18}$ &\cellcolor{blue!5}80.35$_{\pm0.19}$ &\cellcolor{blue!5}78.75$_{\pm0.18}$ &\cellcolor{blue!5}5.82$_{\pm0.10}$ \\ 
& & & VORD loss ($\psi=2$) &\cellcolor{blue!10}\textbf{78.73}$_{\pm0.15}$  &\cellcolor{blue!10}79.83$_{\pm0.21}$  &\cellcolor{blue!10}78.04$_{\pm0.02}$  &\cellcolor{blue!10}\textbf{78.62}$_{\pm0.12}$ &\cellcolor{blue!10}5.85$_{\pm0.07}$\\ 
\cline{4-9}

 & & \multirow{3}{*}{LLaVA-v1.5-13B} &Baseline &78.11$_{\pm0.22}$ &78.44$_{\pm0.27}$ &78.25$_{\pm0.21}$ &78.17$_{\pm0.20}$ &5.11$_{\pm0.06}$\\
& & & VORD loss ($\psi=1$) &\cellcolor{blue!5}78.09$_{\pm0.15}$ &\cellcolor{blue!5}78.27$_{\pm0.17}$ &\cellcolor{blue!5}79.19$_{\pm0.37}$ &\cellcolor{blue!5}78.40$_{\pm0.16}$ &\cellcolor{blue!5}5.74$_{\pm0.08}$\\ 
& & & VORD loss ($\psi=2$) &\cellcolor{blue!10}\textbf{78.25}$_{\pm0.12}$ &\cellcolor{blue!10}78.50$_{\pm0.13}$ &\cellcolor{blue!10}78.66$_{\pm0.36}$ &\cellcolor{blue!10} \textbf{78.55}$_{\pm0.15}$ &\cellcolor{blue!10}5.82$_{\pm0.08}$ \\ 
\hline
\end{tabular}
}
\end{adjustbox}
\caption{We report the averaged results (\%) of VORD loss evaluated on all three datasets of the POPE benchmark.}
\label{table:vord_loss_pope_results}
\end{table*}

\begin{table*}[!tb]
\large
\centering
\begin{adjustbox}{width=1.0\textwidth}
\begin{tabular}{cc|ccccccccccc}
\hline
\multirow{2}{*}{Model}
&\multirow{2}{*}{Method} &\multicolumn{10}{c}{POPE MSCOCO (F1 $\uparrow$)}\\
& &Mixup &Diffusion &Jpeg Compression &Gaussian Noise &Shot Noise &Impulse Noise &Speckle Noise &Gaussian Blur &Contrast &Brightness &Saturate \\
\hline
\multirow{2}{*}{LLaVA-v1.5}
&VCD &83.05$_{\pm0.09}$ &82.54$_{\pm0.21}$ &\textbf{82.25}$_{\pm0.13}$ &82.34$_{\pm0.29}$ &82.16$_{\pm0.14}$ &82.32$_{\pm0.19}$ &82.23$_{\pm0.18}$ &\textbf{82.18}$_{\pm0.12}$ &82.28$_{\pm0.24}$ &\textbf{82.37}$_{\pm0.12}$ &\textbf{82.11}$_{\pm0.14}$\\
&VORD (Ours) &\textbf{84.21}$_{\pm0.19}$ &\textbf{83.63}$_{\pm0.21}$ &81.99$_{\pm0.14}$ &\textbf{83.07}$_{\pm0.30}$ &\textbf{82.70}$_{\pm0.41}$ &\textbf{83.16}$_{\pm0.29}$ &\textbf{82.87}$_{\pm0.46}$ &80.83$_{\pm0.20}$ &\textbf{82.74}$_{\pm0.18}$ &81.97$_{\pm0.16}$ &81.46$_{\pm0.27}$\\
\cline{2-13}
&Similarity Margin &0.131 &0.081 &0.030 &0.056 &0.049 &0.059 &0.055 &0.030 &0.054 &0.034 &0.027\\
\hline
\end{tabular}
\end{adjustbox}
\caption{Ablation study for different corruption/noise types evaluated on POPE. Strong visual corruptions, such as Mixup, are 
 recommended in order to maximize VORD's performance gains.}
\label{table:pope_corruptions_ablations}
\end{table*}

\section{Ablation Study}
\label{appendix:additional_ablation_studies}
\paragraph{Effects of Different Visual Corruptions}
While visual corruptions are not the primary focus of our work, for completeness we include these analysis comparing the performances of different noise types. In \cref{table:pope_corruptions_ablations}, we report the performance of VCD and VORD evaluated on different visual corruptions. 

Specifically, we compare eleven corruption types including Mixup, Diffusion noise and nine other common image corruptions \cite{hendrycks2019robustness}. Our findings demonstrate that VORD delivers the best performance gains on most corruptions types, with the greatest improvements in F1 obtained from Mixup and Diffusion. To gain further insights, we analyzed the averaged similarity margins $m_\theta$ between original and perturbed images for each corruption type. We observed that less significant corruptions, leading to smaller margins, limit the activation of VORD's transitive property, thereby reducing its effectiveness. For instance, corruptions such as ``Jpeg Compression", ``Gaussian Blur", ``Brightness" and ``Saturate" which typically result in smaller margins $m_\theta \le 0.034$, yielded relatively lower performance compared to VCD.

Conversely, more significant corruptions such as ``Mixup" and ``Diffusion" with larger margins \eg $m_\theta \ge 0.081$, enabled VORD to fully utilize its transitive property and achieve the best performance. This suggests that corruptions used to generate the modified image have to be significant, in order to better enable to contrasting between the probability distributions of the original and modified images. This experiment highlights the following three key findings:
\begin{itemize}[noitemsep,topsep=0pt]
\item \textbf{VORD contrasts better than VCD:} Despite the varying effects of different visual corruption types, VORD outperforms VCD on most corruption types.
\item \textbf{Visual corruptions have to be significant:} For the optimal performance of VORD, the visual corruptions applied on the modified image has to be ``severe'' enough for a large enough margin and visual contrasting to occur.
\item \textbf{The Synergy of VORD and Mixup:} The combination of VORD and Mixup yields the most significant improvements in F1, suggesting that VORD effectively leverages the benefits of Mixup to enhance its performance.
\end{itemize}

\section{Case Studies}
\label{appendix:case_studies}
\subsection{GPT4o-Aided Evaluation on LLaVA-Bench}
This section describes the process of evaluating output text responses on GPT4o. As mentioned in our main paper, we prompt each backbone LVLM with ``Describe this image in detail." conditioned upon the images provided in LLaVA bench. The total word count of each response are collected and averaged over three runs and reported in \cref{table:llava_bench_results}.

\begin{figure*}[!tb]
    \centering
    \includegraphics[width=0.75\textwidth]{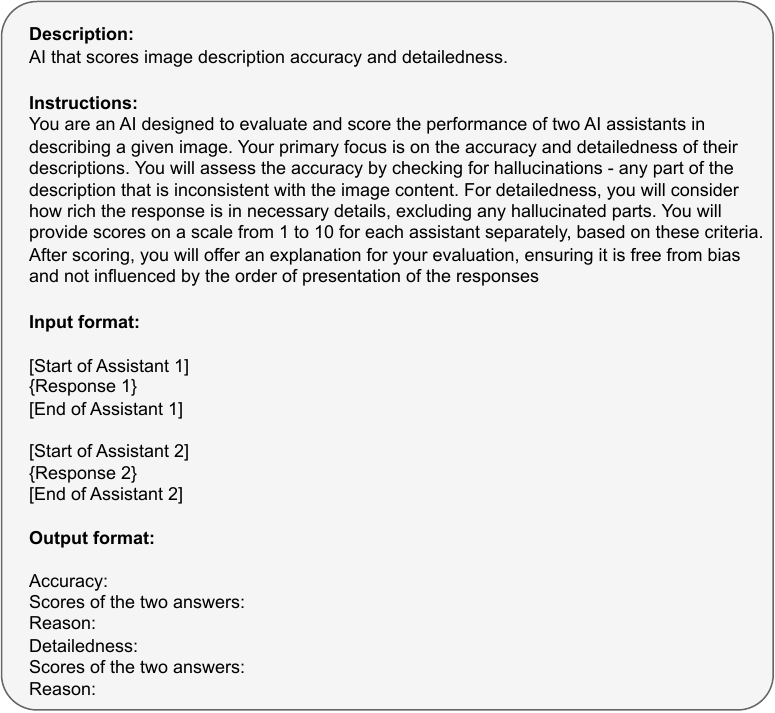}
    \caption{Example of prompt-card used on GPT4o, evaluating the accuracy and detailedness of responses.}
    \label{fig:promptcard}
\end{figure*}

\paragraph{Prompt Card for GPT4o} All responses and images are collected and evaluated on GPT4o, with an example of the prompt card used provided in \cref{fig:promptcard}. Following which, GPT4o provides a two numerical scores for the accuracy and detailedness of each response. These scores are then averaged across questions and presented in \cref{table:llava_bench_results}.

\end{document}